\documentclass[]{aux/bytedance_seed}
\usepackage[toc,page,header]{appendix}
\usepackage{minitoc}
\usepackage{cleveref} 
\usepackage{subcaption}
\usepackage{booktabs}
\usepackage{graphicx}   
\usepackage{subcaption} 
\usepackage{siunitx}    
\usepackage{amsmath}    
\usepackage{adjustbox}
\usepackage{amssymb} 
\usepackage{wrapfig}

\usepackage{amsmath,amsfonts,bm}

\def\eqref#1{equation~\ref{#1}}

\def\1{\bm{1}}

\DeclareMathAlphabet{\mathsfit}{\encodingdefault}{\sfdefault}{m}{sl}
\SetMathAlphabet{\mathsfit}{bold}{\encodingdefault}{\sfdefault}{bx}{n}

\def\gR{{\mathcal{R}}}

\newcommand{\E}{\mathbb{E}}

\usepackage{natbib}
\usepackage{latexsym}

\usepackage{url}
\usepackage{amssymb}
\usepackage[utf8]{inputenc}
\usepackage{microtype}
\usepackage{booktabs}
\usepackage{pifont} 
\usepackage{multirow}
\usepackage{makecell}
\usepackage{paralist}
\usepackage{xspace}
\usepackage{color}
\usepackage{xcolor}
\usepackage{colortbl}
\usepackage{adjustbox}
\usepackage{hyperref} 
\usepackage[edges]{forest}
\usepackage{tikz} 
\usepackage{caption}
\usepackage{amsfonts}

\hypersetup{
    colorlinks,
    linkcolor={blue!80!black},
    citecolor={blue!80!black},
}
\tikzset{
    root/.style =             {align=center, text width=1cm, rounded corners=3pt, line width=0.3mm, fill=gray!10, draw=gray!80, font=\small},
    demographic/.style =         {align=center, text width=1.8cm, rounded corners=3pt, line width=0.3mm, fill=blue!10, draw=blue!80, font=\footnotesize},
    demographic_work/.style =    {align=center, text width=10cm, rounded corners=3pt, line width=0.3mm, fill=blue!10, draw=blue!0, font=\footnotesize},
    character/.style =         {align=center, text width=1.8cm, rounded corners=3pt, line width=0.3mm, fill=red!10, draw=red!80, font=\footnotesize},
    character_work/.style =    {align=center, text width=10cm, rounded corners=3pt, line width=0.3mm, fill=red!10, draw=red!0, font=\footnotesize},
    personalization/.style =           {align=center, text width=1.8cm, rounded corners=3pt, line width=0.3mm, fill=cyan!10, draw=cyan!80, font=\footnotesize},
    personalization_work/.style =      {align=center, text width=10cm, rounded corners=3pt, line width=0.3mm, fill=cyan!10, draw=cyan!0, font=\footnotesize},
    risk/.style =         {align=center, text width=1.8cm, rounded corners=3pt, line width=0.3mm, fill=orange!10, draw=orange!80, font=\footnotesize},
    risk_work/.style =    {align=center, text width=10cm, rounded corners=3pt, line width=0.3mm, fill=orange!10, draw=orange!0, font=\footnotesize},
}

\usepackage{CJK}

\definecolor{ForestGreen}{RGB}{34, 139, 34}
\definecolor{IndianRed}{RGB}{205, 92, 92}
\newcommand{\increase}[2]{{#1} \textbf{\tiny{(\textcolor{ForestGreen}{$\uparrow$#2})}}}
\newcommand{\decrease}[2]{{#1} \textbf{\tiny{(\textcolor{IndianRed}{$\downarrow$#2})}}}

\title{Know When to Explore: Difficulty-Aware Certainty as a Guide for LLM Reinforcement Learning}

\author[1]{Ang Li \protect\textsuperscript{\dag}}
\author[2]{Zhihang Yuan \protect\textsuperscript{\ddag}}
\author[2]{Yang Zhang}
\author[2]{Shouda Liu}
\author[1]{Yisen Wang}

\affiliation[1]{PKU}
\affiliation[2]{ByteDance Seed}

\abstract{
Reinforcement Learning with Verifiable Feedback (RLVF) has become a key technique for enhancing the reasoning abilities of Large Language Models (LLMs). However, its reliance on sparse, outcome-based rewards—which only indicate if a final answer is correct or not—fails to provide granular guidance on the reasoning process itself. This limitation hinders efficient learning, as the model cannot distinguish between high-quality and inefficient solutions, nor can it learn effectively from different types of failures. To address this, we observe that an LLM's self-certainty often correlates with task difficulty and solution quality.
We introduce \textbf{D}ifficulty-\textbf{A}ware \textbf{C}ertainty-guided \textbf{E}xploration (DACE), a novel RL algorithm that leverages this insight to dynamically balance the exploration-exploitation trade-off. DACE assesses task difficulty online based on the policy's success rate. It then uses this signal to modulate an intrinsic reward: for difficult tasks where the model is struggling, DACE encourages exploration by penalizing high certainty; for easier tasks, it encourages learning efficiency by rewarding high certainty. Experiments on challenging mathematical reasoning benchmarks (AIME, MATH) show that DACE significantly outperforms strong baselines. The DACE-trained models not only achieve higher accuracy but also demonstrate more robust performance when scaling test-time compute, validating that our adaptive approach fosters effective exploration without sacrificing precision.
}
\date{\today}

\begin{document}

\maketitle

\begingroup\renewcommand\thefootnote{}\footnotetext{\dag  Work done during internship at ByteDance.}\endgroup

\begingroup\renewcommand\thefootnote{}\footnotetext{\ddag Project lead.}\endgroup

\section{Introduction}

Large Language Models (LLMs) have demonstrated remarkable capabilities in complex reasoning domains such as mathematics and programming \cite{o1, guo2025deepseek}. A key driver of this success has been Reinforcement Learning with Verifiable Feedback (RLVF), a paradigm that fine-tunes models using rewards derived from rule-based verifiers \cite{deepseekmath}. These verifiers offer a scalable and objective method for assigning binary rewards (e.g., 0 for incorrect, 1 for correct). However, this reliance on sparse, outcome-based rewards introduces a significant limitation: the inability to provide granular feedback on the reasoning process itself. A verifier cannot distinguish between two correct solutions of varying quality—one being concise and elegant, the other verbose and inefficient. Similarly, it treats all incorrect solutions as equally uninformative, failing to guide the model away from specific reasoning fallacies.

\begin{figure}[htbp]
    \centering
    \includegraphics[width=1.0\linewidth]{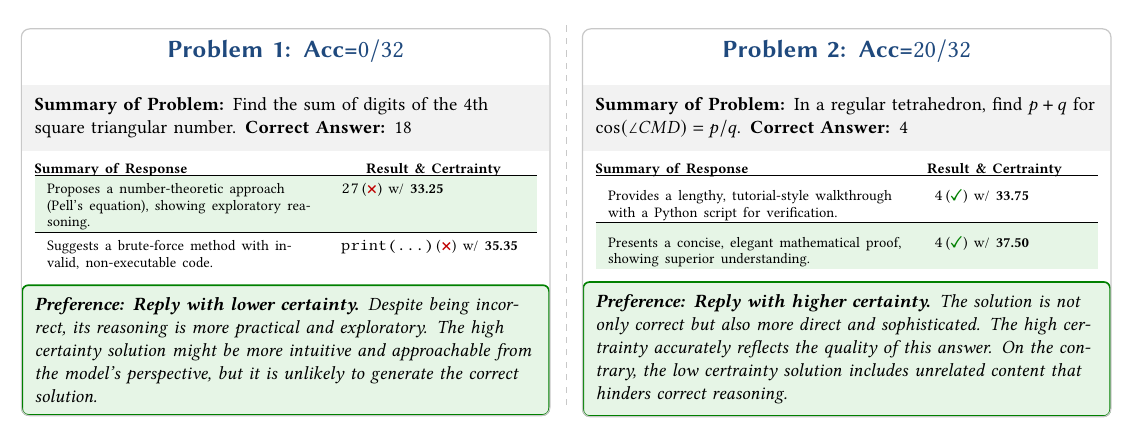}
    \caption{Self-certainty of LLMs can provide meaningful training signals for responses of identical outcome rewards. \textbf{Left:} For a challenging problem, where the model is often wrong when certain, low-certainty responses are more likely to be exploratory and novel. \textbf{Right:} When the model has a high chance of solving a problem, higher certainty often indicates a more direct and concise solution. Both cases are from testing Qwen2.5-7B on the AMC23 benchmark.}
    \label{fig:overall_intro}
\end{figure}

This issue is illustrated in Figure \ref{fig:overall_intro}, which showcases responses from Qwen2.5-7B \cite{yang2024qwen2} to problems from the American Mathematics Competitions benchmark. When generating incorrect solutions (left), the model may exhibit qualitatively different failures; some represent promising but flawed reasoning paths worth exploring, while others are simple dead ends. When generating correct solutions (right), some are notably more direct and efficient. From a policy learning perspective, an ideal reward signal would encourage exploration away from common failure modes while simultaneously steering the policy to exploit and refine efficient, correct solutions. This level of guidance is unattainable with simple binary rewards.

While prior work has explored process-based rewards \cite{lightman2023let,wang2023math,cui2025processreinforcementimplicitrewards}, these methods can be costly to scale. We turn instead to intrinsic signals generated by the model itself. Observing Figure \ref{fig:overall_intro} again, we note that the model's self-certainty—its confidence in its own generation \cite{kang2025scalablebestofnselectionlarge}—correlates with these qualitative differences. For challenging problems, low-certainty responses may signal valuable exploration. For simpler problems, high-certainty responses often reflect efficient, reliable reasoning. This observation leads to our central research question:

\begin{center}
\textit{Can we leverage an intrinsic signal like self-certainty to dynamically guide the model's exploration-exploitation trade-off based on its real-time assessment of task difficulty?}
\end{center}

In this paper, we answer this question affirmatively. We first use a toy RL environment to demonstrate that the optimal strategy is contingent on task difficulty: exploration is vital for hard tasks, while exploitation accelerates convergence on easy ones. Building on this insight, we introduce \textbf{D}ifficulty-\textbf{A}ware \textbf{C}ertainty-guided \textbf{E}xploration (DACE), a novel algorithm that dynamically adjusts its learning objective. For problems the model finds difficult (i.e., has a low success rate), DACE encourages lower certainty to promote exploration. Conversely, for problems the model has mastered, it encourages higher certainty to exploit and refine its knowledge.

Our extensive experiments on large-scale mathematical reasoning benchmarks validate the effectiveness of DACE. Key findings include:
\begin{itemize}
    \item DACE consistently outperforms a strong GRPO baseline, delivering significant performance gains on challenging, competition-level datasets like AIME25 (+1.3) and AIME24 (+2.9).
    \item Ablation studies confirm that DACE's adaptive strategy is superior to fixed strategies of pure exploration or pure exploitation, which are shown to be suboptimal.
    \item The performance advantage of DACE-trained models widens when scaling test-time compute, demonstrating that our method fosters a more robust and diverse set of correct reasoning paths without sacrificing precision.
\end{itemize}

Our work demonstrates that harnessing intrinsic signals to intelligently navigate the exploration-exploitation dilemma is a powerful as well as easy to plug-in approach for LLM training in sparse-reward settings. We believe DACE represents a notable step toward more effective reinforcement learning for complex reasoning.

\section{Related Works}
Our work, DACE, is situated at the intersection of two major research areas: the foundational exploration-exploitation dilemma in Reinforcement Learning (RL) and the emerging use of intrinsic signals for guiding LLM reasoning. We first review the classical RL context before examining how these principles have been adapted, and often simplified, for LLMs, thereby motivating our approach for a more adaptive strategy.

\textbf{The Exploration-Exploitation Dilemma in Reinforcement Learning.}
The need to balance gathering new information (exploration) with leveraging known rewards (exploitation) is a fundamental challenge in RL, particularly in domains with sparse rewards \cite{kayal2025impactintrinsicrewardsexploration}. In the context of LLM reasoning, where a reward may only be granted for a complete and correct final answer, this challenge is especially acute.
Theoretical approaches have established near-optimal regret bounds in structured settings like bandits and MDPs \cite{ThompsonSamplingTheory, agrawal2013thompson, efroni2020explorationexploitationconstrainedmdps, osband2017posterior}. In practical deep RL, this trade-off is managed through various techniques, including lookahead planning \cite{ecoffet2021first, florensa2018automatic}, policy randomization \cite{sutton1998reinforcement, tokic2010adaptive, tang2017exploration, osband2016deep}, and—most relevant to our work—the use of intrinsic rewards. The core idea of intrinsic motivation is to augment the sparse external reward ($r_{ext}$) with a dense, internally-generated signal ($r_{int}$), optimizing for a combined objective:
\begin{align}
\arg\max \E_{\pi}[r_{ext} + \beta r_{int}],
\end{align}
where $\beta$ controls the influence of the intrinsic reward. Researchers have proposed many forms of $r_{int}$ to encourage exploration, such as novelty-based signals derived from state counts \cite{tang2017exploration} or prediction errors (curiosity) \cite{schmidhuber1991curious, pathak2017curiosity, burda2018exploration,sekar2020planning}, information gain \cite{houthooft2016vime}, or skill diversity \cite{eysenbach2018diversity}. A key limitation of many of these methods is that the intrinsic signal is applied uniformly, often with a fixed goal (e.g., always seek novelty), without adapting to the agent's growing competence or the specific difficulty of the current task.

\textbf{Intrinsic Signals for LLM Reasoning.}
When applying these concepts to LLM reasoning, the research has diverged into two distinct streams, creating a dichotomy between purely exploitative and purely explorative objectives.
On one hand, a significant body of work uses intrinsic signals to enforce exploitation. These methods often operate without any external reward ($r_{ext}$) and aim to refine the LLM's existing knowledge by maximizing signals like self-consistency \cite{zuo2025ttrltesttimereinforcementlearning}, self-certainty \cite{zhao2025learningreasonexternalrewards}, or confidence (i.e., negative entropy) \cite{zhang2025rightquestionhalfanswer, agarwal2025unreasonableeffectivenessentropyminimization, gao2025oneshotentropyminimization}. While successful, particularly for test-time adaptation, these approaches are inherently conservative and may struggle with problems that require novel lines of reasoning beyond the model's initial high-confidence paths.
On the other hand, a second stream of work uses intrinsic signals to promote exploration. Here, the primary focus has been on maximizing policy entropy, either to understand its role in reasoning \cite{wang20258020rulehighentropyminority, gao2025oneshotentropyminimization} or to explicitly encourage diverse thought processes during supervised training \cite{zheng2025returnentropyelicitingexplore, cheng2025reasoningexplorationentropyperspective}. Other intrinsic signals include the probability ratio between the current and a reference model, used to guide the generation process \cite{cui2025process, yuan2024free}. These methods are effective at discovering new strategies but risk over-exploring or failing to commit to a promising solution.

Our work, DACE, bridges this gap. We argue that the choice to explore or exploit should not be static. Instead, it should be a dynamic decision guided by the model's awareness of the task's difficulty. We propose using self-certainty not as a fixed goal to be maximized, but as an adaptive intrinsic signal. When certainty is high on an easy problem, DACE encourages exploitation. When certainty is low on a difficult problem, it encourages exploration. In this way, DACE leverages a well-established intrinsic signal \cite{kang2025scalablebestofnselectionlarge} to create a nuanced, difficulty-aware balance between exploration and exploitation for LLM reasoning.

\section{Motivating DACE: The Need for Adaptive Exploration}
\label{sec:motivation}

\begin{figure}[htbp] 
    \centering 
    \begin{subfigure}[b]{0.49\linewidth}
        \centering
        \includegraphics[width=\linewidth]{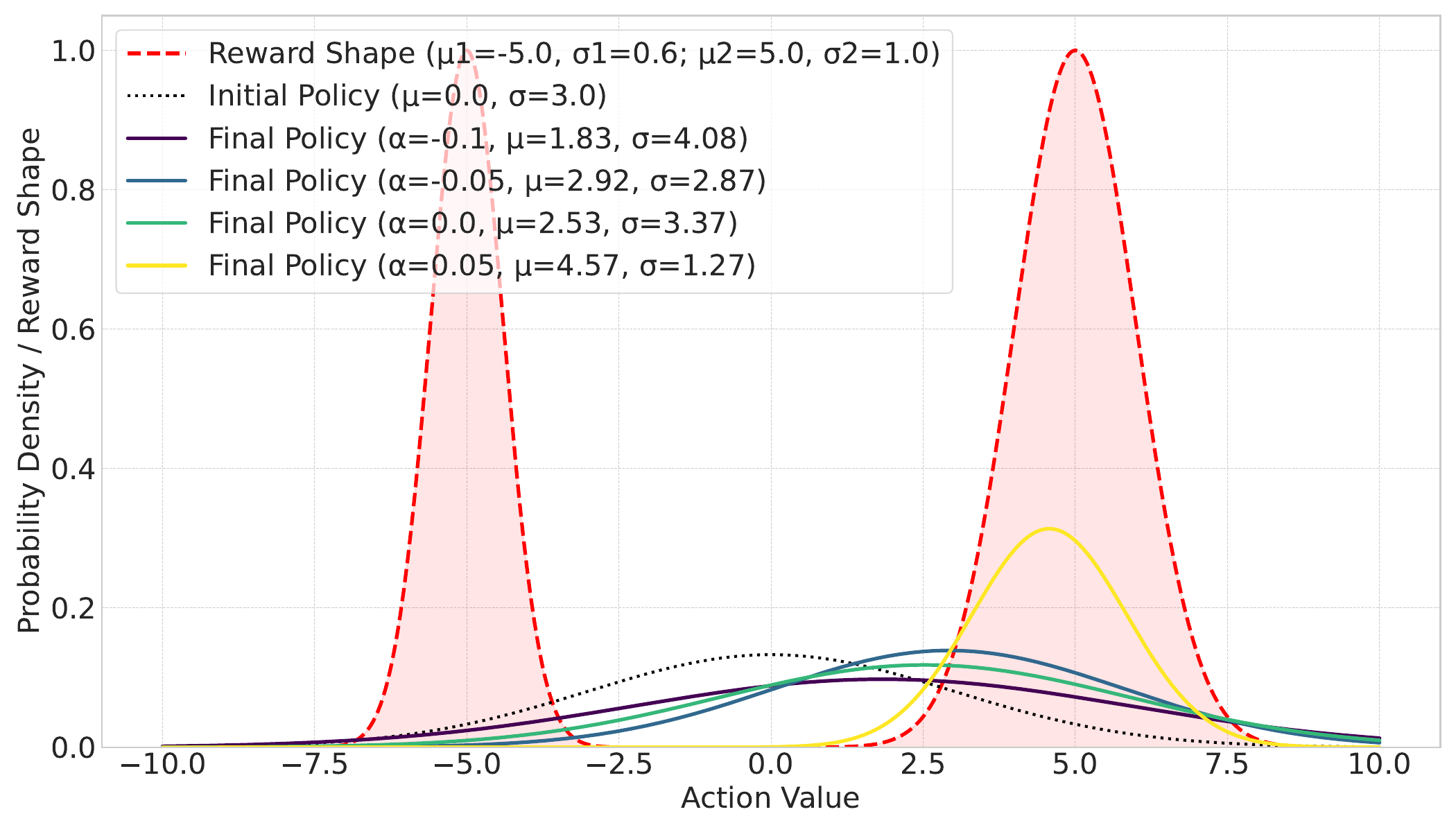}
        \caption{Difficult task (sparse rewards).}
        \label{fig:toy-example-1}
    \end{subfigure}
    \hfill 
    \begin{subfigure}[b]{0.49\linewidth}
        \centering
        \includegraphics[width=\linewidth]{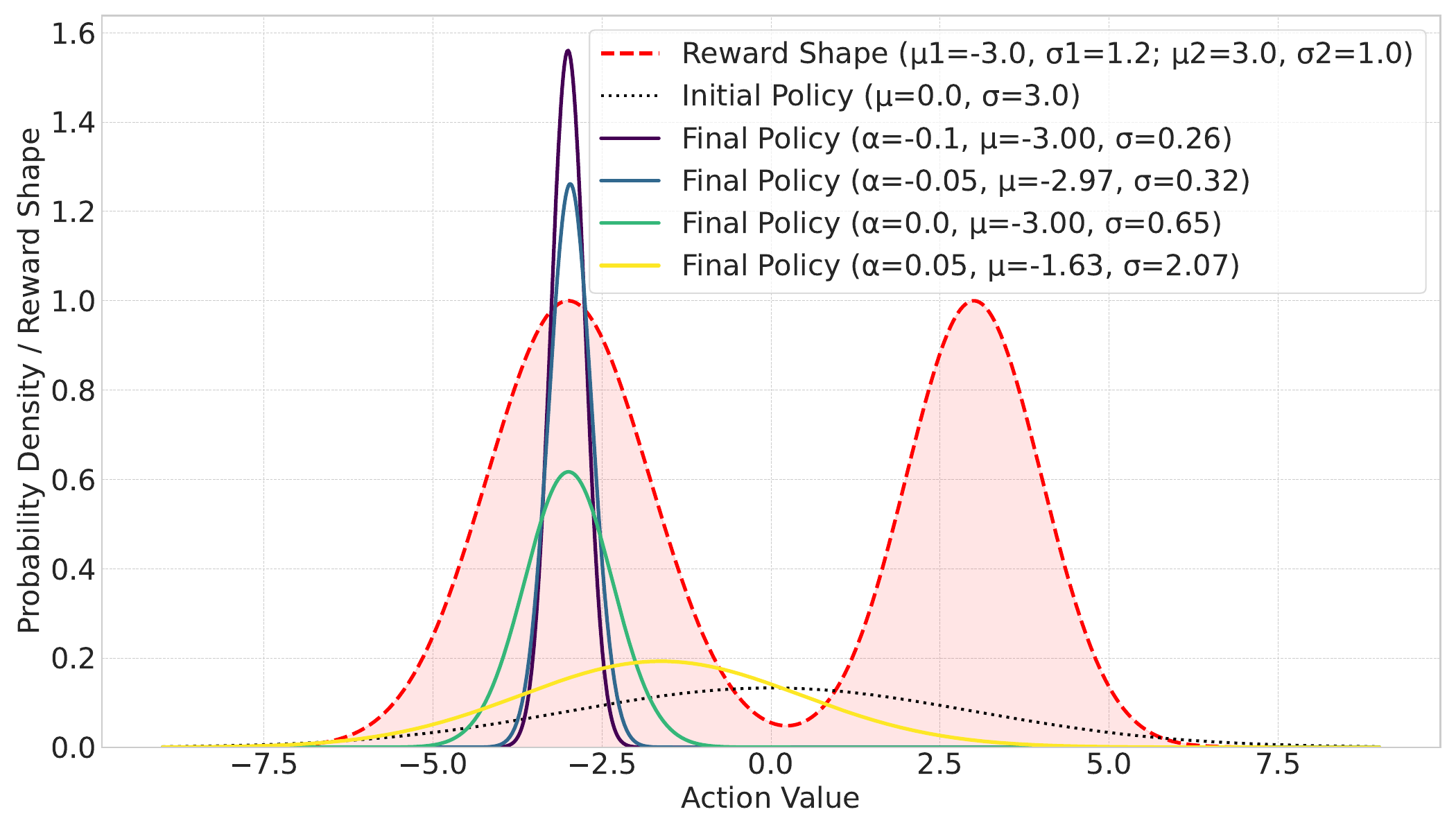}
        \caption{Easy task (dense rewards).}
        \label{fig:toy-example-2}
    \end{subfigure}
    \caption{The optimal certainty strategy depends on task difficulty. \textbf{(a)} On a difficult task, lowering certainty (blue curve, $\alpha>0$) is necessary for exploration and discovering the reward. \textbf{(b)} On an easy task, increasing certainty (purple curve, $\alpha<0$) accelerates convergence via exploitation.}
    \label{fig:toy-example} 
\end{figure}

\par We begin by presenting an illustrative case study to build the intuition for our proposed method, \textbf{D}ifficulty-\textbf{A}ware \textbf{C}ertainty-guided \textbf{E}xploration (DACE). This study demonstrates a core principle: the optimal balance between exploration and exploitation is not static but depends critically on task difficulty. The findings from this simplified setting reveal the limitations of fixed strategies and provide a clear rationale for an adaptive approach like DACE.

\subsection{A Toy Learning Setting}
\par We construct a simple continuous reinforcement learning environment where an agent's policy is a one-dimensional Gaussian distribution, $\pi(a; \mu, \sigma) \sim \frac{1}{\sqrt{2\pi \sigma}} \exp(-\frac{(a-\mu)^2}{2\sigma^2})$. Here, the mean $\mu$ represents the agent's belief about the optimal action, while the standard deviation $\sigma$ represents its certainty. A small $\sigma$ indicates high certainty (exploitation), whereas a large $\sigma$ indicates low certainty (exploration). The reward function is a mixture of two Gaussian distributions, $\gR(a) = \gR(a; -\mu_r, \sigma_r) + \gR(a; \mu_r, 1.0)$, where $\gR(a; \mu_r, \sigma_r) \sim \exp(-\frac{(a-\mu_r)^2}{2\sigma_r^2})$. By adjusting $\mu_r$ and $\sigma_r$, we can simulate environments with varying reward landscapes, from sparse (difficult) to dense (easy).

\subsection{Certainty as a Static Learning Objective}
\par We quantify policy certainty using its log standard deviation, $\ln\sigma$, and incorporate it directly into the learning objective:
\begin{equation}
    \max_{\mu, \sigma} \E_{a\sim \pi(\cdot|\mu, \sigma)} \big[\gR(a) + \alpha \ln\sigma\big],
    \label{eq:toy_objective}
\end{equation}
where the hyperparameter $\alpha$ dictates a fixed learning strategy. We compare three distinct, static configurations:
\begin{itemize}
    \item \textbf{Forced Exploration ($\alpha > 0$):} The policy is always encouraged to increase its variance (lower its certainty), analogous to Maximum Entropy RL.
    \item \textbf{Standard RL ($\alpha = 0$):} The policy only maximizes the expected reward, with no explicit control over certainty.
    \item \textbf{Forced Exploitation ($\alpha < 0$):} The policy is always penalized for high variance, encouraging it to shrink its distribution and exploit known rewards.
\end{itemize}
We use Proximal Policy Optimization (PPO) to optimize the policy under these fixed strategies. Further experimental details are in Appendix \ref{sec:appen_toy_model_details}.

\begin{table}[htbp]
    \centering
    \caption{Quantitative evaluation of fixed certainty strategies. The table reports the final expected reward for different $\alpha$ values under varying task difficulties (controlled by $\sigma_r^1$). The highest reward for each difficulty is in \textbf{bold}. The optimal $\alpha$ transitions from positive (exploration) to negative (exploitation) as the task becomes easier (larger $\sigma_r^1$).}
    \label{tab:toy_mean_reward_analysis}
    \begin{tabular}{
        l 
        S[table-format=1.2e-2] S[table-format=1.3e-2] S[table-format=1.4] S[table-format=1.4] S[table-format=1.3] S[table-format=1.3] S[table-format=1.3]
    }
        \toprule
        & \multicolumn{7}{c}{\textbf{Task Difficulty (Width of Reward Distribution $\sigma_r^1$)}} \\
        \cmidrule(lr){2-8}
        \textbf{$\alpha$} & {\textbf{0.6}} & {\textbf{0.7}} & {\textbf{0.8}} & {\textbf{0.9}} & {\textbf{1.0}} & {\textbf{1.1}} & {\textbf{1.2}} \\
        \midrule
        -0.1  & 3.19e-22 & 1.92e-16 & 0.00177 & 0.0405 & \textbf{0.289} & \textbf{0.797} & \textbf{0.949} \\
        -0.05 & 3.57e-22 & 1.11e-14 & 0.00441 & 0.0573 & 0.285 & 0.611 & 0.926 \\
        0.0   & 4.40e-22 & 3.46e-16 & 0.00411 & 0.0776 & 0.233 & 0.523 & 0.878 \\
        0.05  & \textbf{5.34e-22} & \textbf{0.000911} & \textbf{0.0106} & \textbf{0.0972} & 0.218 & 0.404 & 0.755 \\
        \bottomrule
    \end{tabular}
\end{table}

\subsection{Observation: Optimal Strategy Depends on Difficulty}
\par Our experiments reveal a clear dependency between the optimal learning strategy and the difficulty of the task, which we analyze both qualitatively and quantitatively.

\textbf{Qualitative Analysis.} Figure \ref{fig:toy-example} illustrates two contrasting scenarios. In a difficult setting with sparse rewards (Figure \ref{fig:toy-example-1}), a fixed exploration strategy ($\alpha > 0$) is superior. By lowering its certainty, the policy explores a wider action space and successfully discovers the distant reward mode. Conversely, in an easier setting with dense rewards (Figure \ref{fig:toy-example-2}), a fixed exploitation strategy ($\alpha < 0$) converges much faster. By increasing its certainty, the policy quickly hones in on the obvious optimal action.

\textbf{Quantitative Analysis.} We confirm this finding quantitatively in Table \ref{tab:toy_mean_reward_analysis}. We vary the task difficulty by adjusting the width of one reward peak, $\sigma_r^1$, from narrow (harder) to wide (easier). The results show a clear transition point. For difficult tasks with sparse rewards ($\sigma_r^1 \le 0.9$), an explorative strategy ($\alpha > 0$) achieves the highest final reward. As the task becomes easier with denser rewards ($\sigma_r^1 \ge 1.0$), an exploitative strategy ($\alpha < 0$) becomes dominant.

\par Taken together, these findings highlight a critical limitation in static learning strategies: no single choice of $\alpha$ is optimal across all scenarios. This motivates the central idea behind DACE: an agent should not be forced to always explore or always exploit, but should instead learn to \textit{adapt} its strategy based on its own assessment of the task's difficulty.

\section{DACE: Difficulty-Aware Certainty-guided Exploration}
\label{sec:method}

\par Building on the insight from our case study, we introduce \textbf{D}ifficulty-\textbf{A}ware \textbf{C}ertainty-guided \textbf{E}xploration (DACE). DACE is an RL algorithm designed to dynamically balance the exploration-exploitation trade-off. It achieves this by operationalizing the key insight from our motivating example: the agent must first assess the difficulty of a given task and then use that assessment to modulate its own policy certainty, deciding whether to explore or exploit. DACE is composed of three core components: difficulty estimation, a certainty metric, and a mechanism to connect them.

\textbf{Difficulty-Awareness:}
The first component of DACE is a mechanism to quantify task difficulty. Crucially, difficulty is not an absolute property of a task but is relative to the policy's current capabilities. We therefore define an online difficulty measure for a query $x$ with respect to the current policy $\pi$ as its estimated failure rate. This is a practical, policy-relative proxy for difficulty, which we approximate by sampling $n$ responses and averaging their outcomes from a binary verifier:
\begin{equation}
    \text{diff}(x;\pi) = 1 - \E_{y\sim \pi(\cdot| x)}\I \big[\text{verify}(y) = 1\big] \approx 1 - \frac{1}{n} \sum_{i=1}^n \I[\text{verify}(y_i) = 1].
    \label{eq:difficulty}
\end{equation}
A higher $\text{diff}(x;\pi)$ value indicates a problem that the current policy finds more challenging.

\textbf{Certainty as a Lever for Exploration.}
The second component is a way to measure and control the policy's behavior. Generalizing from the standard deviation ($\sigma$) in our toy example, we define policy certainty for autoregressive LLMs based on prior work \cite{kang2025scalablebestofnselectionlarge}. The certainty of a generated sequence $y$ is its negative average log-probability:
\begin{equation}
    C(y, x; \pi) = -\frac{1}{|y|} \sum_{j=1}^{|y|} \log \pi(y_j | x, y_{<j}).
    \label{eq:certainty}
\end{equation}
This metric serves as a lever to control behavior. Maximizing certainty encourages the policy to use high-probability tokens, leading to deterministic, \textit{exploitative} behavior. Conversely, minimizing certainty allows for lower-probability tokens, promoting diverse and \textit{explorative} behavior.

\textbf{Connecting Difficulty to Certainty via an Adaptive Intrinsic Reward.}
The core of DACE is the intrinsic reward that forges a dynamic link between policy-relative difficulty (Eq.~\ref{eq:difficulty}) and behavioral certainty (Eq.~\ref{eq:certainty}). We achieve this with an adaptive coefficient, $\alpha(x; \pi)$, that modulates the certainty-based reward:
\begin{equation}
    R_{\text{int}}(x, y; \pi) = \alpha(x; \pi) \cdot C(y, x; \pi),
    \label{eq:intrinsic_reward}
\end{equation}
where the coefficient's sign is determined by comparing the task difficulty to a threshold:
\begin{equation}
    \alpha(x; \pi) = \alpha_{\text{scale}} \cdot \text{sgn}\left(\beta_{\text{threshold}} - \text{diff}(x; \pi)\right).
    \label{eq:alpha_adaptive}
\end{equation}
Here, $\alpha_{\text{scale}} > 0$ is a scaling factor, and $\beta_{\text{threshold}}$ is a difficulty threshold. This formulation creates the desired adaptive behavior:
\begin{itemize}
    \item \textbf{For hard tasks ($\text{diff}(x; \pi) > \beta_{\text{threshold}}$):} The coefficient $\alpha(x; \pi)$ becomes negative. The objective becomes maximizing $-\alpha_{\text{scale}} \cdot C(y, x; \pi)$, which is equivalent to \textit{minimizing} policy certainty. DACE thus encourages the agent to \textbf{explore} when it is struggling.
    \item \textbf{For easy tasks ($\text{diff}(x; \pi) < \beta_{\text{threshold}}$):} The coefficient $\alpha(x; \pi)$ is positive. The objective encourages maximizing certainty. DACE thus pushes the policy to \textbf{exploit} and refine its successful strategies on problems it can already solve.
\end{itemize}

\textbf{The Full DACE Objective.}
The complete DACE learning objective integrates this adaptive intrinsic reward with the standard external reward from the environment:
\begin{equation}
    \max_{\pi} \; \E_{x\sim D,\, y \sim \pi(\cdot | x)} \left[ R_{\text{ext}}(x, y) + \alpha(x; \pi) \, C(y, x ; \pi) \right],
    \label{eq:full_objective}
\end{equation}
where $R_{ext}$ is the external reward function (e.g., accuracy) that we want to maximize. This objective dynamically adjusts the learning pressure based on the agent's real-time performance on a given task. We optimize this objective using Group-wise Rejection Policy Optimization (GRPO). This choice is highly synergistic, as the $n$ samples required by GRPO for its policy update can be directly reused for the difficulty estimation in Equation~\ref{eq:difficulty}. This makes the implementation of DACE both elegant and computationally efficient.


\section{Experiments}
\label{sec:experiments}
We now empirically evaluate \textbf{D}ifficulty-\textbf{A}ware \textbf{C}ertainty-guided \textbf{E}xploration (DACE) to validate our central hypothesis: that dynamically balancing exploration and exploitation based on task difficulty enhances the mathematical reasoning capabilities of LLMs.

\textbf{Setup.} We use Qwen2.5-7B \citep{yang2024qwen2} as our base model and conduct RL training within the VeRL framework \cite{sheng2024hybridflow}. Our training data is a de-duplicated mixture of the DAPO \cite{yu2025dapoopensourcellmreinforcement} and MATH \cite{hendrycks2021measuring} datasets. For all runs, we adopt the clip-higher trick from DAPO with $\epsilon_{low}=0.2$ and $\epsilon_{high}=0.28$. For DACE, our default configuration uses a scaling factor of $\alpha_{\text{scale}}=0.05$ and a difficulty threshold of $\beta_\text{threshold}=0.4$. We set the group size for our policy optimizer to $16$ and omit standard entropy/KL regularization terms. A complete list of hyper-parameters is provided in Appendix \ref{sec:appen_RL_training_details}.

\textbf{Evaluation.} We follow standard protocols for assessing mathematical reasoning, reporting mean@32 accuracy on four widely-used benchmarks: AIME25, AIME24, AMC23, and MATH-500 \cite{hendrycks2021measuring}. For evaluation, we use a temperature of $0.6$ and top-k of $30$.

\textbf{Baselines.} We compare DACE against a strong GRPO baseline \cite{deepseekmath} with the clip-higher trick, as well as several advanced methods. These include \textit{Ent-Adv} \citep{cheng2025reasoningexplorationentropyperspective}, which promotes exploration by encouraging longer reasoning; \textit{Clip-Cov} and \textit{KL-Cov} \citep{cui2025entropymechanismreinforcementlearning}, which control updates based on token-entropy covariance; and \textit{FR3E} \citep{zheng2025returnentropyelicitingexplore}, which performs targeted rollouts on high-entropy tokens. Due to the high cost of reproduction, we report the results for these baselines directly from their original papers.

\subsection{Main Results}
The empirical results, summarized in Table \ref{tab:main_result}, demonstrate the effectiveness of DACE's adaptive strategy. Our method consistently outperforms the strong GRPO baseline on the most challenging benchmarks.

\begin{table}[htbp]
    \centering
    \caption{Performance comparison on mathematical reasoning benchmarks. DACE consistently improves over the GRPO baseline and achieves state-of-the-art results on AIME25 and AMC23. $^*$ denotes results reported in the original papers. The highest score for each benchmark is marked in \textbf{bold}.}
    \label{tab:main_result}
    \renewcommand{\arraystretch}{1.2} 
    \begin{tabular}{ | l | c | c | c | c |}
        \toprule
         \textbf{Datasets} & AIME25 & AIME24 & AMC23 & MATH-500 \\
         \midrule
         \multicolumn{5}{| c |}{\textit{Qwen2.5-7B}} \\
         \midrule
         Base & 2.2 & 5.2 & 28.3 & 54.4 \\
         GRPO & 15.9 & 14.6 & 70.9 & \textbf{82.1} \\
         \midrule
         ~~~w/ Ent-Adv. $^*$ & 11.8 & 12.6 & 57.8 & 58.5 \\
         ~~~w/ Clip-Cov $^*$ & 15.8 & 22.1 & 58.2 & 80.4 \\
         ~~~w/ KL-Cov $^*$ & 12.9 & 22.6 & 61.4 & 80.8 \\
         ~~~w/ FR3E $^*$ & -- & \textbf{25.2} & 67.5 & 79.0 \\
         \midrule
         ~~~w/ DACE (ours) & 
         \textbf{\increase{17.2}{+1.3}}  &
         \increase{17.5}{+2.9}  &
         \textbf{\increase{71.4}{+0.5}}  &
         \decrease{81.9}{-0.2} \\
         \bottomrule
    \end{tabular}
\end{table}

Specifically, DACE achieves absolute gains of \textbf{+1.3} points on AIME25, \textbf{+2.9} on AIME24, and \textbf{+0.5} on AMC23 over GRPO. These improvements establish DACE as the top-performing method on both AIME25 and AMC23 among all listed approaches. While FR3E leads on AIME24, DACE still shows a significant improvement over GRPO and other techniques. On MATH-500, DACE's performance is on par with the strong GRPO baseline, showing only a marginal 0.2-point difference. These results suggest that DACE's dynamic approach—exploring on hard problems while exploiting on easy ones—is particularly effective for the most complex, competition-level reasoning tasks.

\begin{wrapfigure}{r}{0.48\textwidth}
    \centering
    \vspace{-10pt}
    \includegraphics[width=1.0\linewidth]{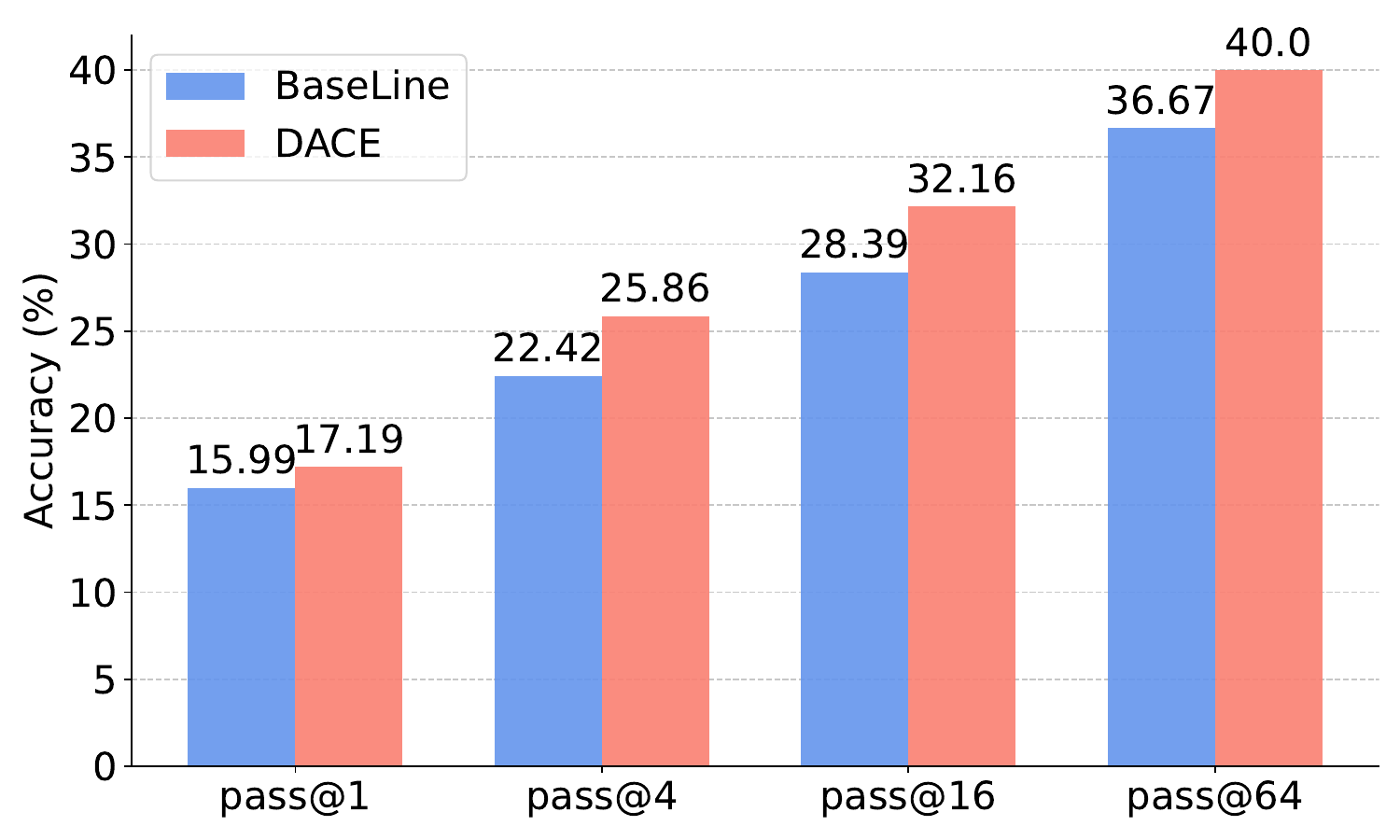} 
    \caption{Impact of scaling test-time compute on the AIME25 benchmark (pass@k). The DACE-trained model consistently outperforms the GRPO baseline, and the performance gap widens as more responses are sampled.}
    \vspace{-10pt}
    \label{fig:scaling_test_compute}
\end{wrapfigure}

\par \textbf{Scaling Test-Time Compute.} We next investigate how models trained with DACE perform when more computational resources are allocated at test time. As shown in Figure \ref{fig:scaling_test_compute}, the DACE-trained model maintains a consistent and significant performance advantage over the GRPO baseline on AIME25. Notably, the performance gap widens as we increase the number of samples, growing from a +1.2 point lead at mean@16 to a +3.3 point lead at mean@128. This outcome validates the core hypothesis of DACE. By encouraging exploitation (higher certainty) on easier problems, the model maintains high precision for low sample counts (pass@1). Simultaneously, by encouraging exploration (lower certainty) on harder problems, it discovers a wider range of correct solutions, boosting performance at high sample counts (pass@k). This adaptive strategy effectively fosters robust exploration without sacrificing precision.

\begin{figure}[htbp]
    \centering
    \begin{subfigure}[b]{0.32\linewidth}
        \includegraphics[width=\linewidth]{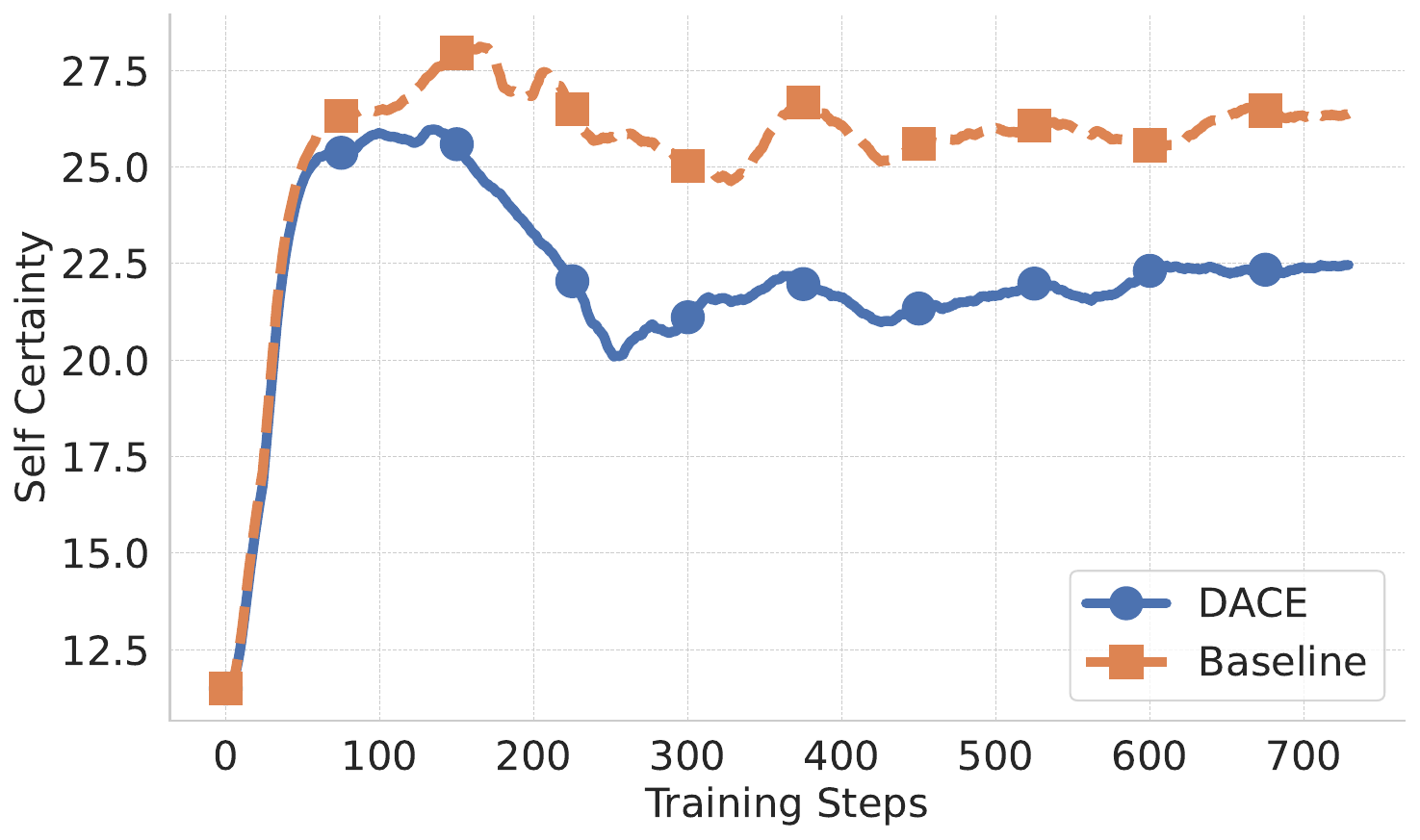}
        \caption{Self-Certainty}
        \label{fig:self-certainty}
    \end{subfigure}
    \hfill
    \begin{subfigure}[b]{0.32\linewidth}
        \includegraphics[width=\linewidth]{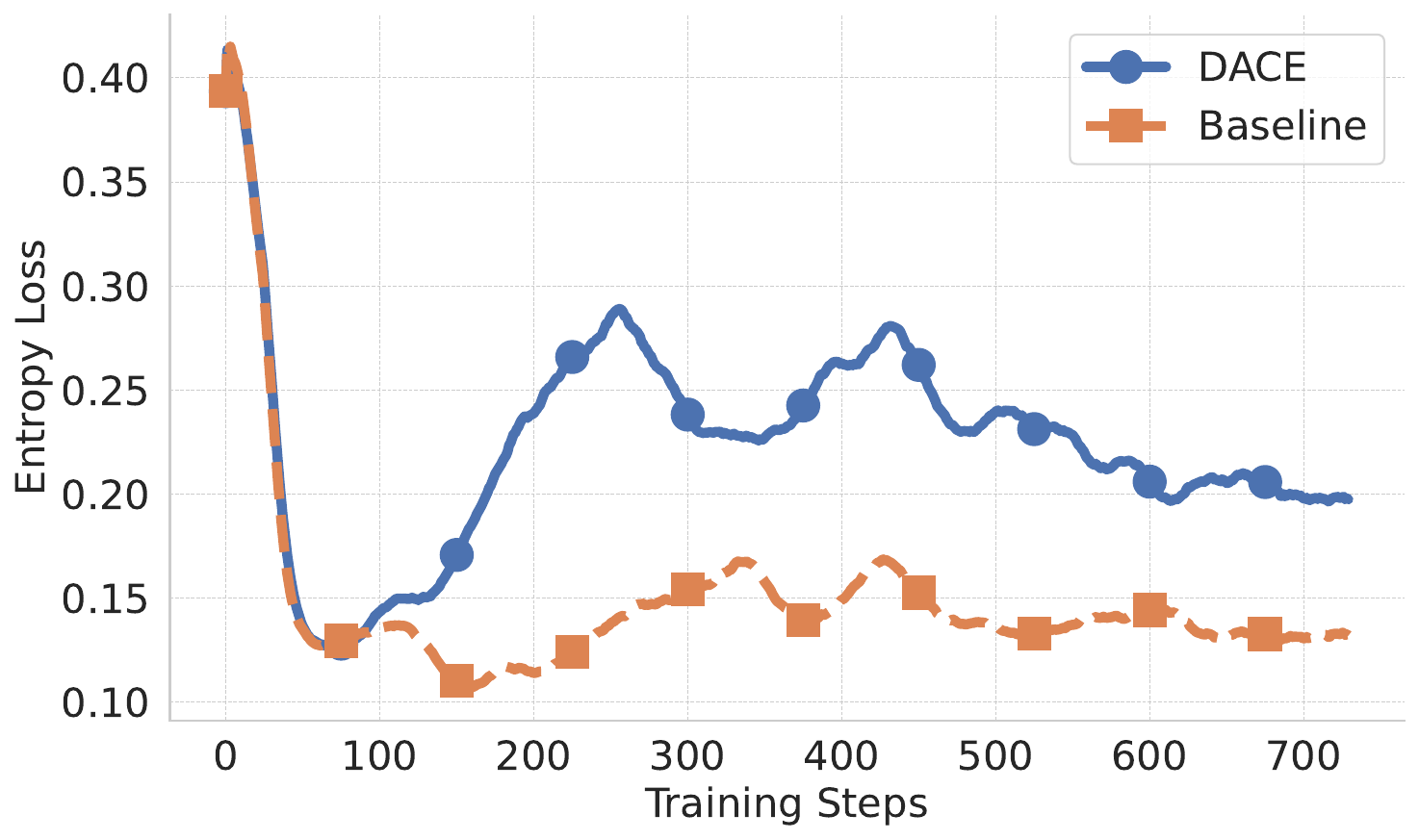}
        \caption{Entropy}
        \label{fig:entropy}
    \end{subfigure}
    \hfill
    \begin{subfigure}[b]{0.32\linewidth}
        \includegraphics[width=\linewidth]{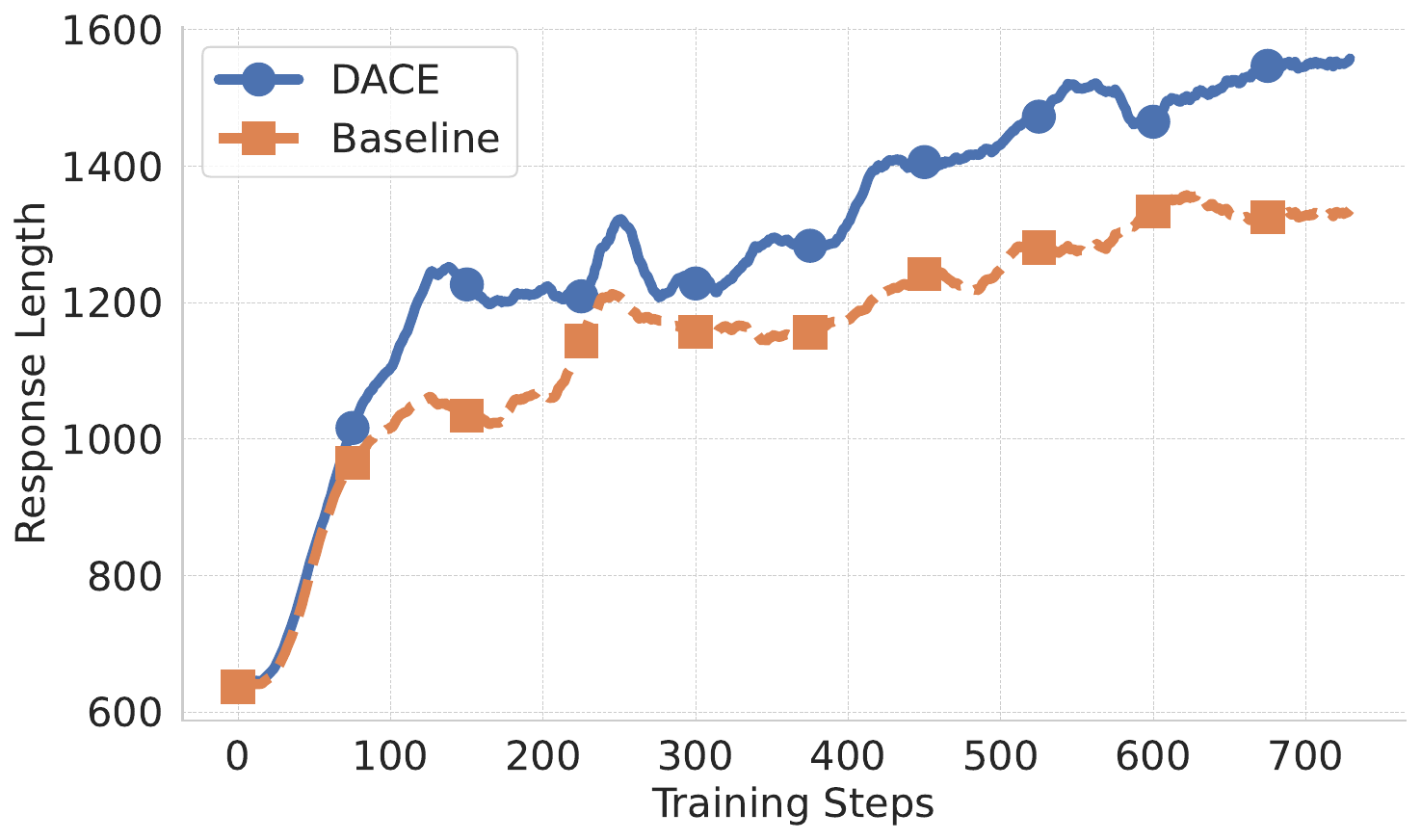}
        \caption{Response Length}
        \label{fig:exp-response-length}
    \end{subfigure}
    \caption{Training dynamics comparison between DACE and GRPO. The plots show the evolution of (a) actor self-certainty, (b) token-level entropy, and (c) mean response length. DACE exhibits a more exploratory pattern, characterized by lower certainty and higher entropy, particularly in the intermediate training stages.}
    \label{fig:training_dynamic_baseline_compare}
\end{figure}

\par \textbf{Training Dynamics.} An analysis of the training process reveals the mechanism behind DACE's performance gains. Figure \ref{fig:training_dynamic_baseline_compare} shows that DACE's adaptive reward induces a distinctly more exploratory training behavior compared to GRPO. This is evidenced by consistently lower model self-certainty, higher token-level entropy, and slightly longer responses. Interestingly, the divergence between the two methods is most prominent during the intermediate stages of training before the metrics begin to converge. This suggests that DACE injects a critical phase of exploration mid-training, allowing the model to discover more diverse and robust reasoning strategies that lead to its final performance advantage.

\par \textbf{Summary of Main Results.} Our experiments provide comprehensive evidence for DACE's effectiveness. First, on standard benchmarks, DACE delivers substantial performance gains over a strong GRPO baseline, especially on difficult math problems. Second, this advantage is amplified when scaling test-time compute, demonstrating the robustness of the learned policy. Finally, training dynamics reveal that DACE successfully encourages a more exploratory behavior during critical training phases. This allows the model to discover better reasoning paths without sacrificing precision, confirming that DACE's principle of difficulty-aware certainty guidance is a potent method for enhancing complex reasoning.

\subsection{Understanding DACE: The Role of the Difficulty Threshold}

We now delve deeper into DACE's core mechanism by analyzing the trade-off between exploration and exploitation as we vary the difficulty threshold, $\beta_\text{threshold}$.

\textbf{Setup.} Using our default configuration, we perform a grid search over $\beta_\text{threshold} \in \{0, 0.2, 0.4, 0.6, 0.8, 1.0\}$. The endpoints represent fixed, non-adaptive strategies. A threshold of $\beta_\text{threshold}=0.0$ forces DACE to always treat problems as 'easy', thus always rewarding high certainty (pure exploitation). Conversely, $\beta_\text{threshold}=1.0$ forces DACE to always treat problems as 'hard', thus always penalizing high certainty and promoting maximum entropy (pure exploration).

\begin{figure}[htbp]
    \centering
    \begin{subfigure}[b]{0.49\linewidth}
        \includegraphics[width=\linewidth]{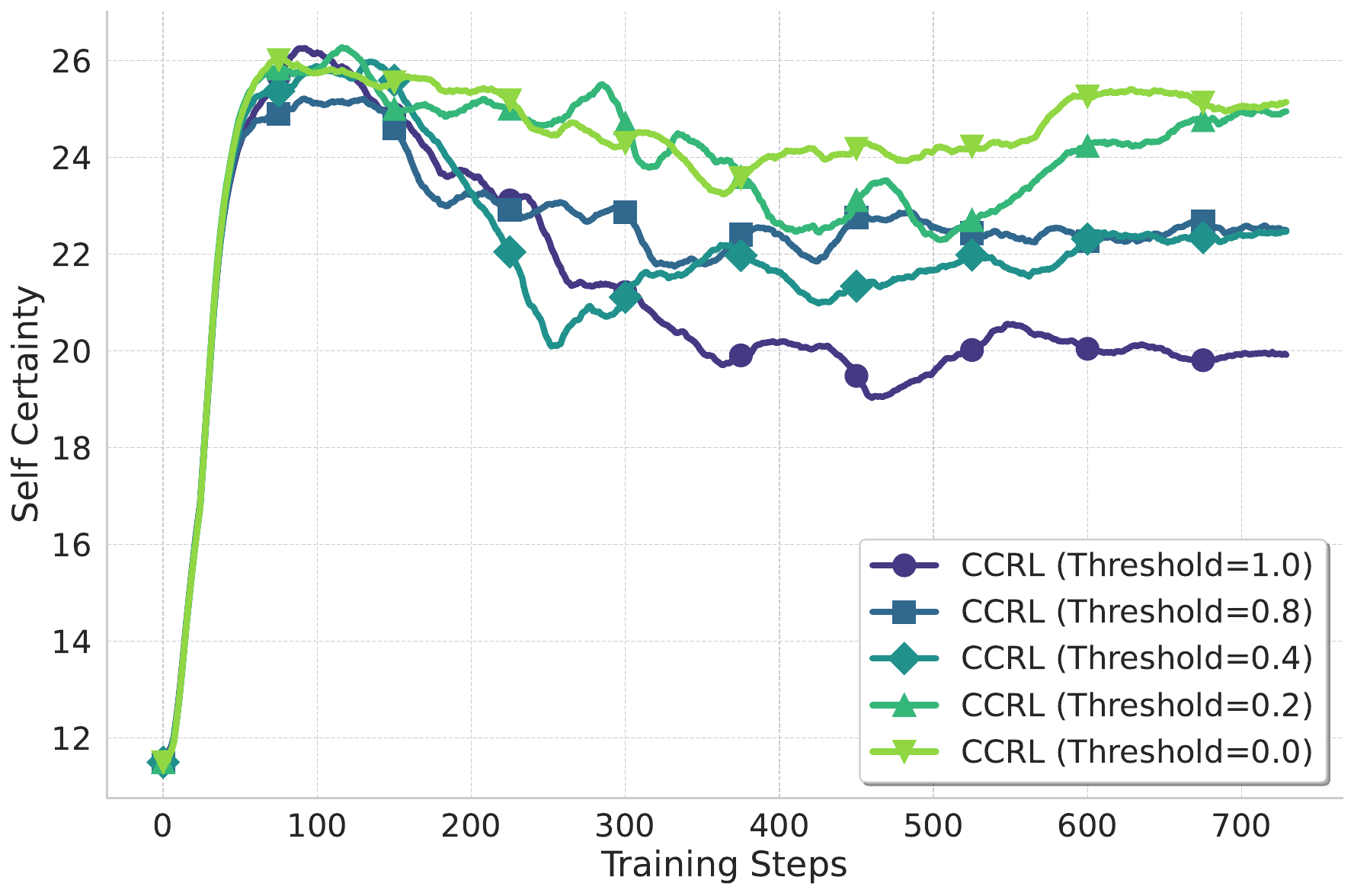}
        \caption{Self-Certainty}
        \label{fig:self-certainty-5runs}
    \end{subfigure}
    \hfill
    \begin{subfigure}[b]{0.49\linewidth}
        \includegraphics[width=\linewidth]{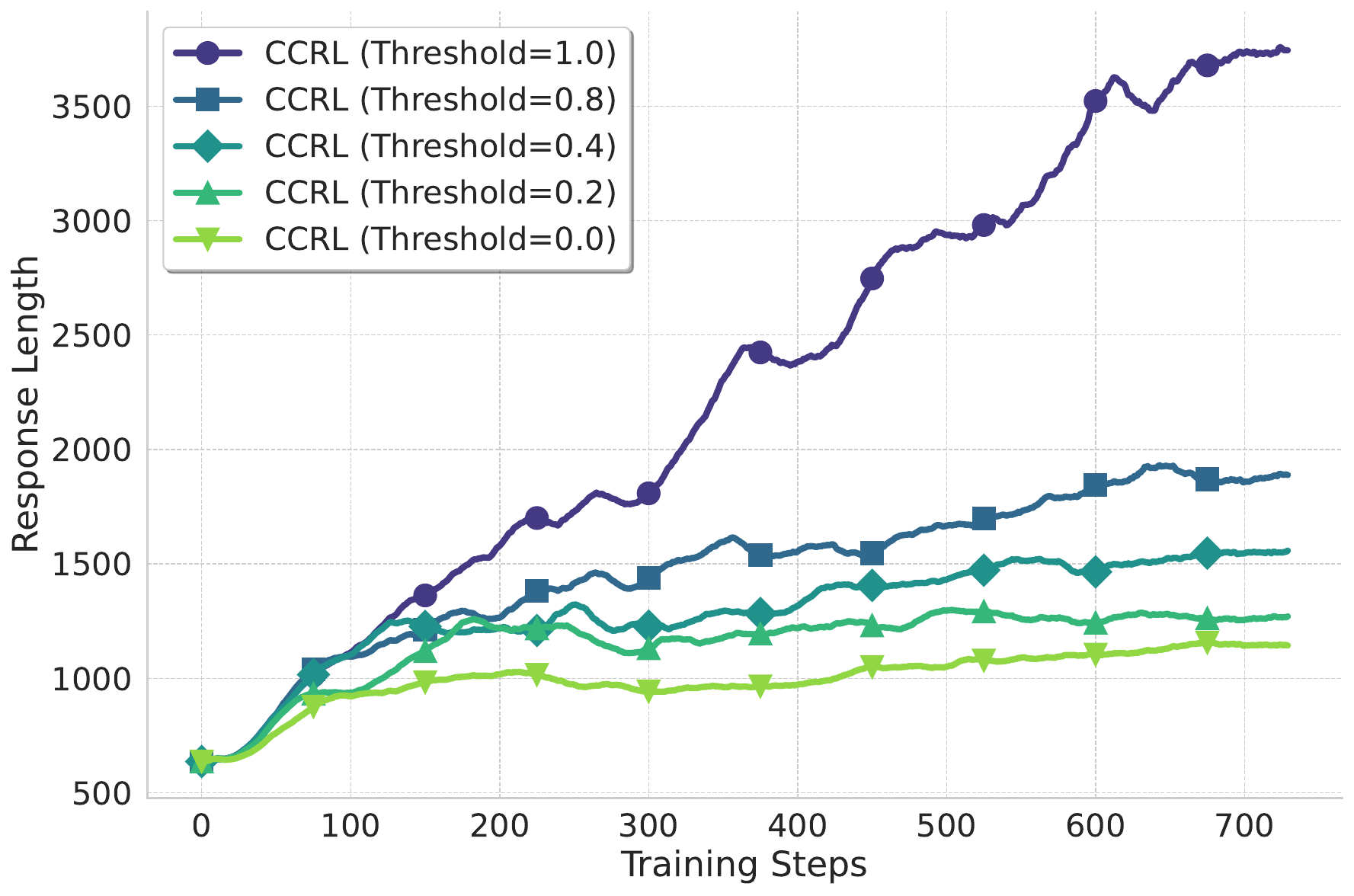}
        \caption{Response Length}
        \label{fig:response-length-5runs}
    \end{subfigure}
    \caption{Impact of the difficulty threshold $\beta_{threshold}$ on training dynamics. $\beta_{threshold}=0.6$ is omitted for clarity. Moving from pure exploitation ($\beta=0.0$) to pure exploration ($\beta=1.0$) systematically decreases policy certainty and increases response length.}
    \label{fig:5runs-training-dynamics}
\end{figure}

\par We first examine how $\beta_\text{threshold}$ influences training dynamics, shown in Figure \ref{fig:5runs-training-dynamics}. As the threshold increases, we observe a clear trend of decreasing self-certainty and increasing response length. This is expected: a higher threshold means more problems are classified as 'hard', triggering the exploratory, certainty-penalizing reward. Notably, the pure exploration strategy ($\beta_\text{threshold}=1.0$) is an outlier, generating responses ~3x longer than other settings. This highlights a known pitfall of constant exploration: it can lead to inefficient, redundant reasoning and poor convergence \cite{cheng2025reasoningexplorationentropyperspective}. In contrast, the pure exploitation strategy ($\beta_\text{threshold}=0.0$) produces the shortest responses, but as we will see, this comes at the cost of accuracy.

\begin{table}[htbp]
    \centering
    \caption{Influence of the difficulty threshold ($\beta_\text{threshold}$) and reward weight ($\alpha_\text{scale}$) on final accuracy (\%).}
    \label{tab:influence}
    \begin{subtable}{\textwidth}
        \centering
        \sisetup{table-format=2.2, round-mode=places, round-precision=2}
        \begin{tabular}{@{}l S[table-format=2.2] S[table-format=2.2] S[table-format=2.2] S[table-format=2.2] S[table-format=2.2, table-space-text-post=\%]@{}}
            \toprule
            \textbf{Threshold ($\beta$)} & {AIME25} & {AIME24} & {AMC23} & {MATH500} & {\textbf{Average}} \\
            \midrule
            0.0 & 15.36 & 14.11 & 70.76 & 81.51 & 45.44 \\
            0.2 & 14.22 & 13.70 & 71.23 & 81.93 & 45.27 \\
            0.4 & 17.19 & 17.45 & 71.37 & 81.56 & 46.89 \\
            0.6 & 14.17 & 15.36 & 74.49 & 81.31 & 46.33 \\
            0.8 & 16.41 & 20.68 & 71.42 & 81.88 & {\bfseries 47.60} \\
            1.0 & 16.21 & 15.58 & 70.31 & 84.60 & 46.68 \\
            \bottomrule
        \end{tabular}
        \caption{Accuracy (\%) with different difficulty thresholds ($\alpha_\text{scale}$ fixed at 0.05).}
        \label{tab:ablate_difficulty}
    \end{subtable}
    \vspace{2.5em}
    \begin{subtable}{\textwidth}
        \centering
        \sisetup{table-format=2.2, round-mode=places, round-precision=2}
        \begin{tabular}{@{}l S[table-format=2.2] S[table-format=2.2] S[table-format=2.2] S[table-format=2.2] S[table-format=2.2, table-space-text-post=\%]@{}}
            \toprule
            \textbf{Weight ($\alpha$)} & {AIME25} & {AIME24} & {AMC23} & {MATH500} & {\textbf{Average}} \\
            \midrule
            0.05 & 17.19 & 17.45 & 71.37 & 81.56 & 46.89 \\
            0.10 & 14.69 & 18.91 & 72.63 & 82.44 & {\bfseries 47.17} \\
            \bottomrule
        \end{tabular}
        \caption{Accuracy (\%) with different weights ($\beta_\text{threshold}$ fixed at 0.4).}
        \label{tab:abalte_weight}
    \end{subtable}
    \vspace{-10mm}
\end{table}

\par Next, we evaluate the impact of $\beta_\text{threshold}$ on final accuracy in Table \ref{tab:ablate_difficulty}. The results clearly show that fixed strategies are suboptimal. Pure exploitation ($\beta=0.0, 0.2$) yields low average accuracy, suggesting the model gets stuck in local optima and fails to discover better reasoning paths. Pure exploration ($\beta=1.0$) is also suboptimal, suffering from inconsistent performance and high computational cost (as seen in response length). The best performance is achieved with intermediate thresholds of $0.4$ and $0.8$, with $\beta_\text{threshold}=0.8$ achieving the highest average accuracy of {47.60\%}. This confirms that the power of DACE lies not in forcing one behavior, but in its ability to \textit{dynamically switch} between them based on task difficulty.

\par Finally, we briefly investigate the influence of the reward scaling factor, $\alpha_\text{scale}$. As shown in Table \ref{tab:abalte_weight}, when fixing the threshold at our default of 0.4, increasing the weight from 0.05 to 0.10 further improves the average accuracy to {47.17\%}. This suggests that tuning the strength of DACE's guidance signal is another promising avenue for optimization.

\section{Conclusion}

In this work, we addressed the challenge of sparse rewards in Reinforcement Learning with Verifiable Feedback (RLVF) for LLM reasoning. We proposed that an LLM's intrinsic self-certainty, when guided by task difficulty, can provide a powerful, granular training signal. We introduced \textbf{D}ifficulty-\textbf{A}ware \textbf{C}ertainty-guided \textbf{E}xploration (DACE), a novel algorithm that dynamically balances the exploration-exploitation trade-off. By assessing its own success rate on a problem, DACE adaptively encourages exploration (lower certainty) on difficult tasks and exploitation (higher certainty) on easier ones. Our experiments on challenging mathematical reasoning benchmarks demonstrated that DACE significantly outperforms strong baselines, and its performance advantage widens as test-time compute is scaled, confirming that it learns a more robust set of solutions.

While DACE proves effective, its reliance on a fixed difficulty threshold presents an opportunity for future work, such as adjusting this parameter in a explicit curriculum learning way \cite{bengio2009curriculum}. Further research could also explore more sophisticated proxies for difficulty and certainty to refine the guidance signal. Besides, intrinsic signals are prone to over-optimization, and we discuss our methods in tackling the challenge in Appendix \ref{appen:reward_hacking}, which we believe is another important research topic in future. Nevertheless, our work validates a key principle: harnessing a model's internal state to intelligently guide its learning strategy is a potent and sample-efficient approach. DACE represents a principled step toward developing more capable agents capable of mastering complex cognitive tasks.

\clearpage

\bibliographystyle{unsrt}
\bibliography{main}

\clearpage

\beginappendix

\section{Additional Experimental Details}

\subsection{Details of the Toy Model Experiment}
\label{sec:appen_toy_model_details}
\par The motivating experiment presented in Section \ref{sec:motivation} was conducted using a custom PPO implementation. The key hyperparameters used for this experiment are detailed in Table \ref{tab:toy_hyper_param}. The standard entropy bonus was disabled, as the exploration-exploitation balance was explicitly controlled by the $\alpha$ coefficient in our objective function (Equation \ref{eq:toy_objective}).

\begin{table}[htbp]
    \centering
    \caption{Hyperparameter configurations for the toy model experiment.}
    \label{tab:toy_hyper_param}
    \begin{tabular}{@{}lc@{}}
        \toprule
        \textbf{Hyperparameter} & \textbf{Value} \\
        \midrule
        Learning Rate ($\eta$) & 0.01 \\
        PPO Clipping Epsilon ($\epsilon$) & 0.2 \\
        Epochs per Update & 10 \\
        Batch Size & 64 \\
        Total Training Iterations & 34 \\
        Steps per Iteration & 32 \\
        \bottomrule
    \end{tabular}
\end{table}

\subsection{Details of RL Training for LLMs}
\label{sec:appen_RL_training_details}
\par All of our large-scale reinforcement learning experiments were conducted using the VeRL framework \cite{sheng2024hybridflow}.
We employed a learning rate schedule with a linear warm-up phase over the first 25 steps, followed by a cosine decay to zero for the remainder of training. The key hyperparameters for our RL training runs are provided in Table \ref{tab:train_hyper_param}.

\begin{table}[htbp]
    \centering
    \caption{Hyperparameter configurations for RL training.}
    \label{tab:train_hyper_param}
    \begin{tabular}{@{}lc@{}}
        \toprule
        \textbf{Hyperparameter} & \textbf{Value} \\
        \midrule
        Sampling Temperature & 0.6 \\
        Max Generation Length & 8192 \\
        Training Epochs & 20 \\
        Learning Rate & 1e-6 \\
        Group Size ($n$) & 16 \\
        Global Batch Size & 512 \\
        PPO Minibatch Size & 32 \\
        \bottomrule
    \end{tabular}
\end{table}

\subsection{Details of Evaluation}
\par For all evaluations, we used the vLLM inference engine \cite{kwon2023efficient} to generate responses from the trained models. The complete set of sampling parameters is listed in Table \ref{tab:infer_hyper_param}. To verify correctness, we first extracted the final answer enclosed within the `\arraybackslash boxed\{...\}` command using regular expressions. We then used the Math-Verify library \cite{Kydlicek_Math-Verify_Math_Verification_2025_misc} to programmatically check the correctness of the extracted answer against the ground truth.

\begin{table}[htbp]
    \centering
    \caption{Hyperparameter configurations for evaluation sampling.}
    \label{tab:infer_hyper_param}
    \begin{tabular}{@{}lc@{}}
        \toprule
        \textbf{Hyperparameter} & \textbf{Value} \\
        \midrule
        Sampling Temperature & 0.6 \\
        Max Generation Length & 8192 \\
        Min-p Sampling & 0.95 \\
        Top-k Sampling & 30 \\
        \bottomrule
    \end{tabular}
\end{table}

\subsection{Mitigating Reward Hacking} 
\label{appen:reward_hacking}
\par During our initial experiments, we observed that the intrinsic certainty signal was prone to reward hacking. To ensure training stability and prevent the model from exploiting the reward function, we implemented two key mitigation strategies.

\textbf{Certainty Normalization.} To prevent the model from simply outputting extreme log-probabilities, we normalized the raw certainty values within each group of $n$ responses. Specifically, we applied a group-wise z-score normalization followed by a min-max scaling to bound the final certainty values within the range $[0, 1]$.

\textbf{Penalizing Code Execution.} We identified a specific failure mode where the model would generate solutions containing Python code snippets, effectively using an implicit, unverified computational tool to arrive at an answer. We classified this behavior as a form of reward hacking against the outcome-based verifier. To discourage this, we assigned a reward of zero to any generated solution that contained executable code.

\end{document}